\documentclass[runningheads]{llncs}
\usepackage{graphicx}
\usepackage{amsmath}
\usepackage{amssymb}
\usepackage{xcolor}
\usepackage{array}
\usepackage{float}

\newcommand{\PreserveBackslash}[1]{\let\temp=\\#1\let\\=\temp}
\newcolumntype{C}[1]{>{\PreserveBackslash\centering}p{#1}}
\begin{document}
\title{A Robust Pose Transformational GAN for Pose Guided Person Image Synthesis}
\titlerunning{A Robust Pose Transformational GAN}
%
\author{Arnab Karmakar\orcidID{0000-0001-7538-3921} \and
Deepak Mishra}
\authorrunning{A. Karmakar et al.}
%
\institute{Indian Institute of Space Science and Technology, Trivandrum, Kerala 695547, India
\email{arnabkarmakar.001@gmail.com, deepak.mishra@iist.ac.in}}
\maketitle              
\begin{abstract}
Generating photorealistic images of human subjects in any unseen pose have crucial applications in generating a complete appearance model of the subject. However, from a computer vision perspective, this task becomes significantly challenging due to the inability of modelling the data distribution conditioned on pose. Existing works use a complicated pose transformation model with various additional features such as foreground segmentation, human body parsing etc. to achieve robustness that leads to computational overhead. In this work, we propose a simple yet effective pose transformation GAN by utilizing the Residual Learning method without any additional feature learning to generate a given human image in any arbitrary pose. Using effective data augmentation techniques and cleverly tuning the model, we achieve robustness in terms of illumination, occlusion, distortion and scale. We present a detailed study, both qualitative and quantitative, to demonstrate the superiority of our model over the existing methods on two large datasets.
\end{abstract}
\section{Introduction}
Given an image of a person, a pose transformation model aims to reconstruct the person's appearance in another pose. While for humans it is very easy to imagine how a person would appear in a different body pose, it has been a difficult problem in computer vision to generate photorealistic images conditioned only on pose; given a single 2D image of the human subject. The idea of pose transformation can help construct a viewpoint invariant representation. This has several interesting applications in 3D reconstruction, movie making, motion prediction or human computer interaction etc.

The task of pose transformation given a single image and a desired pose, is achieved by any machine learning model in basically two steps: (1) learning the significant visual features of the person-of-interest along with the background from the given image, and (2) imposing the desired pose on the person-of-interest, generate a photorealistic image while preserving the previously learned features. Generative Adversarial Networks (GAN) \cite{goodfellow-gan} have been widely popular in this field due to its sharp image generation capability. While the majority of successful pose transformation models use different variation of GANs as their primary component, they give little importance to efficient data augmentation and utilization of inherent CNN features to achieve robustness. Recent developments in this field have been targeted to develop complex deep neural network models with the use of multiple external features such as human body parsing \cite{soft-gated}, semantic segmentation \cite{soft-gated}\cite{balakrishnan}, spatial transformation \cite{balakrishnan}\cite{deformable-gans} etc. Although this is helpful in some scenarios, there is accuracy issues and computational overhead due to each intermediate step that affects the final result.

In this work, we aim to develop an improved end-to-end model for pose transformation given only the input image and the desired pose, and without any other external features. We make use of the Residual learning strategy \cite{resnet} in our GAN architecture by incorporating a number of residual blocks. We achieve robustness in terms of occlusion, scale, illumination and distortion by using efficient data augmentation techniques and utilizing inherent CNN features. Our results in two large datasets, a low-resolution person re-identification dataset Market-1501 \cite{market-1501} and high-resolution fashion dataset DeepFashion \cite{deepfashion} have been demonstrated. Our contributions are two folds: First, we develop an improved pose transformation model to synthesize photorealistic images of a person in any desired pose, given a single instance of the person’s image, and without any external features. Second, we achieve robustness in terms of occlusion, scale and illumination by efficient data augmentation techniques and utilizing inherent CNN features.
\section{Related work}
There has been a lot of research in the field of generative image modelling using deep learning techniques. One line of work follow the idea of Variational Autoencoders (VAE) \cite{kingma-vae} which uses the reparameterization trick to maximize the lower bound of data likelihood \cite{mypaper}. VAEs have been popular for its image interpolation capability, but the generated images lack sharpness and high frequency details. GAN \cite{goodfellow-gan} models make use of adversarial training for generating images from random noise. Most works in pose guided person image generation make use of GANs because of its capability to produce fine details.

Amongst the large number of successful GAN architectures, many were developed upon the DCGAN \cite{dcgan} model  that combines Convolutional Neural Network (CNN) with GANs. Pix2pix \cite{pix2pix} proposed a conditional adversarial network (CGAN) for image-to-image translation by learning the mapping from condition image to target image. Yan et al. \cite{skeleton-aided} explored this idea for pose conditioned video generation, where the human images are generated based on skeleton poses. GANs with different variations of U-Net \cite{unet} have been extensively used for pose guided image generation. The PG$^2$ \cite{pg2} model proposes a 2-step process with a U-Net-like network to generate an initial coarse image of the person conditioned on the target pose and then refines the result based upon the difference map. Balakrishnan et al. \cite{balakrishnan} uses separate foreground and background synthesis using a spatial transformer network and U-Net based generator. Ma et al. \cite{disentangled-pig} uses pose sampling using a GAN coupled with an encoder-decoder model. Dong et al. \cite{soft-gated} produces state-of-the-art results in pose driven human image generation and uses human body parsing as an additional attribute for Warping-GAN rendering. These additional attribute learning generates an overhead in computational capability and affects the final results. Other significant works for pose transfer in the field of person re-identification~\cite{pn-gan}~\cite{fd-gan} mostly deals with low resolution images and a complex training procedure. In this work, we propose a simplified end-to-end model for pose transformation without using additional feature learning at any stage.
\section{Methodology}\label{methodology}
\vspace{-5mm}
\begin{figure}
	\centering
	\includegraphics[width=\linewidth]{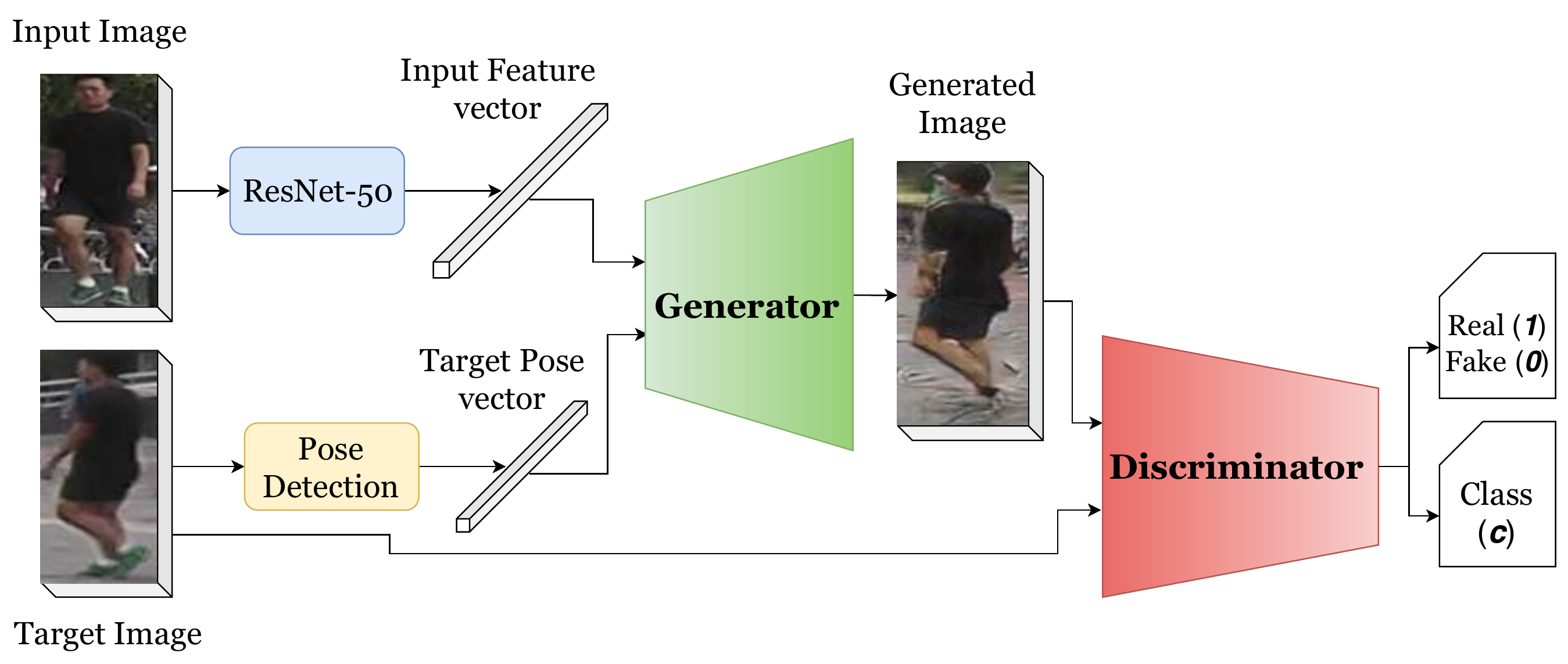}
	\caption{Proposed architecture of the pose transformational GAN (pt-GAN). The idea is to transform the given person image to the desired pose. The additional classification branch of the Discrminator helps the Generator's learning to produce realistic images.}
	\label{fig:gan_arch}
\end{figure}
\vspace{-5mm}
\subsection{Pose Estimation}
The image generation is conditioned on an input image and the target pose represented by a pose vector. In order to get the encoded pose vector, we use off-the-shelf pose detection algorithm OpenPose~\cite{openpose}, which is trained without using either of the datasets deployed in this work. Given an input person image $I_i$, the pose estimation network OpenPose produces a pose vector $P_i$, which localizes and detects 25 anatomical key-points.
\subsection{Generator}
\begin{figure}
	\centering
	\includegraphics[width=0.9\linewidth]{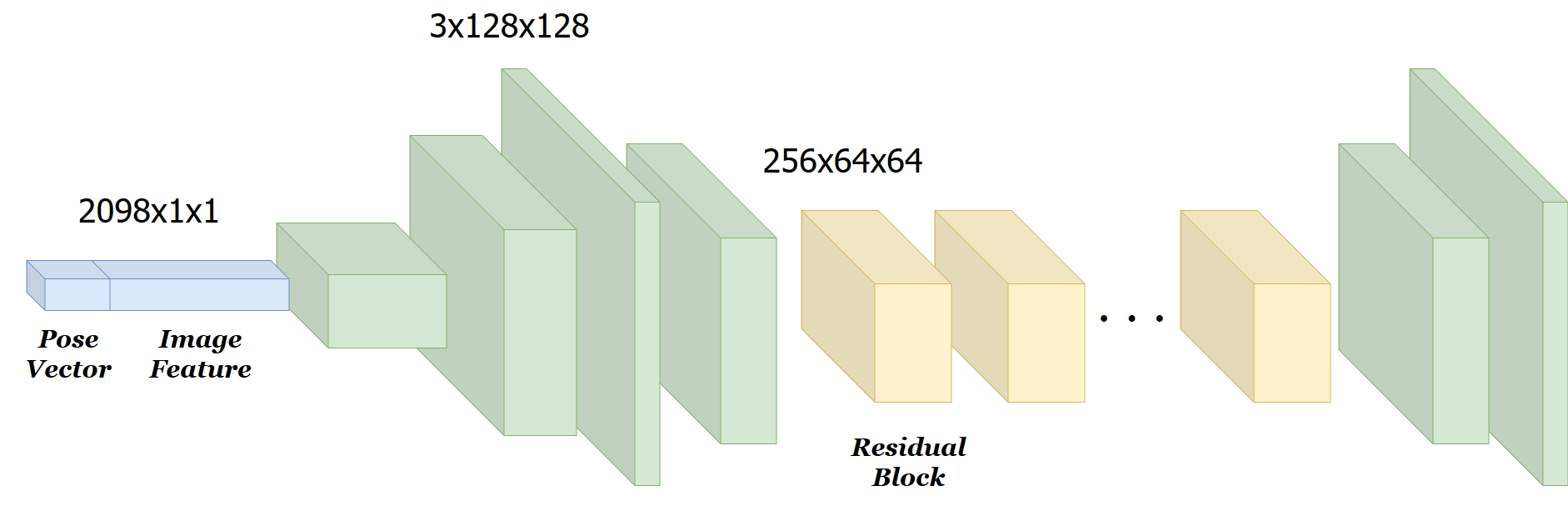}
	\caption{Architecture of the Generator Network. The Generator network consists of 9 residual blocks, which helps the GAN to preserve low level features (clothing, texture), while transforming high level features (pose) of the subject.}
	\label{fig:generator}
\end{figure}

The Image generator ($G_P$) aims at producing the same person's images under different poses. Particularly, given an input person image $I_i$ and a desired pose $P_j$, the generator aims to synthesize a new person image $I_{P_j}$, which contains the same identity but with a different pose defined by $P_j$. The image vector obtained using a pretrained ResNet-50~\cite{resnet} model (ImageNet~\cite{imagenet}), and the pose vector are concatenated and fed to the generator. The architecture is depicted in Figure~\ref{fig:generator}.

The generator consists of multiple Convolution and Transposed Convolution layers. The key element of the proposed Generator is the residual blocks. Each residual block performs downsampling using convolution followed by upsampling using transposed convolution and then re-using the input by addition (Figure \ref{fig:both}(b)).
The motivation is to take advantage of Residual Learning ($y = F(x)+x$) that can be used to pass invariable information (e.g. clothing color, texture, background) from the bottom layers to the higher layers and change the variable information (pose) to synthesize more realistic images, achieving pose transformation at the same time.
\vspace{-5mm}
\subsection{Discriminator}
\begin{figure}
	\centering
	\includegraphics[width=\linewidth]{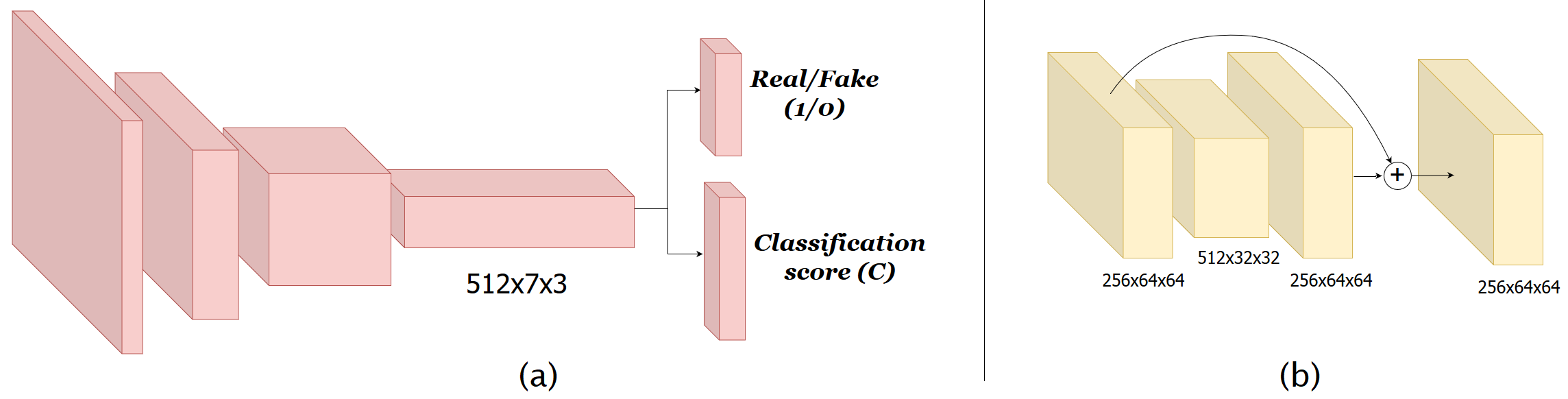}
	\caption{\textbf{(a)} Architecture of the discriminator of pt-GAN. A classification task is added with the real/fake prediction. This simultaneously helps the Generator to produce more realistic images. \textbf{(b)} Architecture of the Residual Blocks used in the Generator. The Residual Learning strategy preserves low level features (color, texture) and learns high level features (pose) simultaneously.}
	\label{fig:both}
\end{figure}

In our implementation, the Discriminator ($D_P$) predicts the class label for the image along with the binary classification of determing whether the image is real or generated. Studies \cite{discrim} show that incorporating classification loss in discriminator along with the real/fake loss, in turn increases the generator's capability to produce sharp images with high details. The Discriminator consists of stacked Conv-ReLU-Pool layer and the final fully connected layer has been modified to incorporate both binary loss and classification loss (Figure~\ref{fig:both}(a)).
\subsection{Data Augmentation}
\begin{enumerate}
	\item \textbf{Image Interpolation}: The input images have been resized to $256\times 256$ before passing through ResNet. Market-1501 images ($128\times 64$) are resized to $256\times 128$, and zero-padded to make $256\times 256$. The images in DeepFashion are of the desired dimension by default.
	\item \textbf{Random Erasing~\cite{randomerasing}}: Random erasing is helpful in achieving robustness against occlusion. A random patch of the input image is given random values while the reconstruction is expected to be perfect. Thus, the GAN learns to reconstruct (and remove) the occluded regions in the generated images.
	\item \textbf{Random Crop}: The input image is randomly cropped and upscaled to the input dimension ($256\times 256$) to augment the cases where the human detection is inaccurate or only the partial body is visible.
	\item \textbf{Jitter}: We use random jitter in terms of brightness, Contrast, Hue and Saturation (random jitter to each channel) to augment the effects of illumination variations.
	\item \textbf{Random Horizontal Flip}: Inspecting the dataset, it is seen that most human subjects has both left-right profile images. Hence flipping the image left-right is a good choice for image augmentation.
	\item \textbf{Random distortion}: We have incorporated random distortion with a grid size of $10$, to compensate the distortion in the generated image as well as enforce our model to learn important features of the input image even in the presence of non-idealities. 
\end{enumerate}
\begin{figure}[H]
	\centering
	\includegraphics[width=0.7\linewidth, clip]{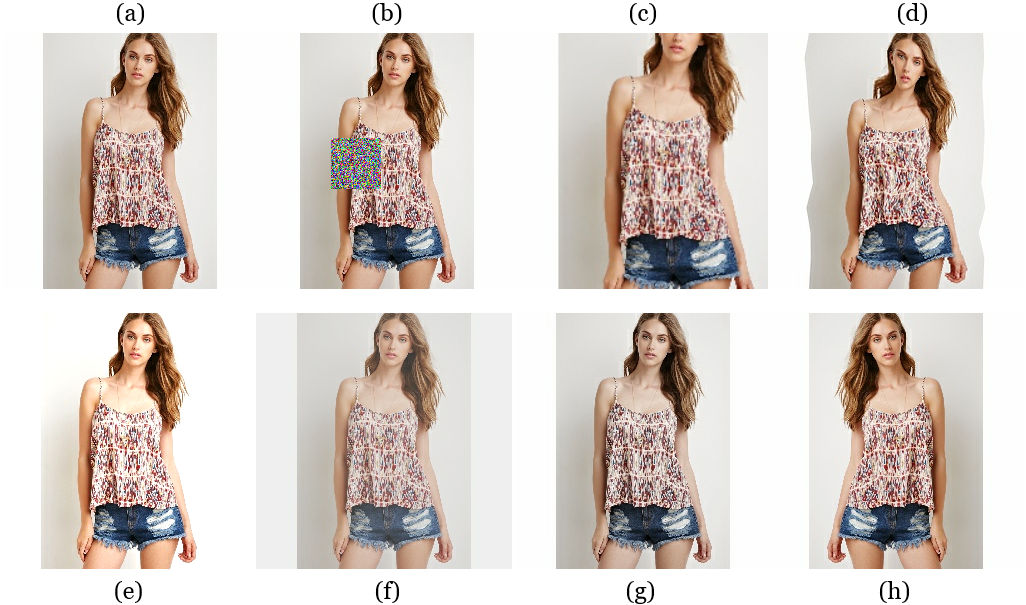}
	\caption{The data augmentation techniques used in this work: (a) Original Image, (b) Random Erasing, (c) Random Crop, (d) Random Distortion, (e)-(g) Random Jitter: (e) Brightness, (f) Contrast, (g) Saturation; (h) Random Flip}
	\label{fig:data_augmentation}
\end{figure}
The CNN by itself enforces scale invariance through max-pooling and convolution layers. Thereafter we claim to have achieved invariance from distortion, occlusion, illumination and scale. A demonstration of the data augmentation techniques is shown in Figure~\ref{fig:data_augmentation}.
\section{Experiments}
\subsection{Datasets}
	\subsubsection{DeepFashion:} The DeepFashion (In-shop Clothes Retrieval Benchmark) dataset \cite{deepfashion} consists of 52,712 in-shop clothes images, and 200,000 cross-pose/scale pairs. The images are of 256$\times$256 resolution. We follow the standard split adopted by \cite{pg2} to construct the training set of 146,680 pairs each composed of two images of the same person but different poses.\vspace{-3mm}
	\subsubsection{Market-1501:} We also show our results on the re-identification dataset Market-1501~\cite{market-1501} containing 32,668 images of 1,501 persons. The images vary highly in pose, illumination, viewpoint and background in this dataset, which makes the person image generation task more challenging. Images have size 128$\times$64. Again, we follow \cite{pg2} to construct the training set of 439,420 pairs, each composed of two images of the same person but different poses.
\subsection{Implementation and Training}
For image descriptor generation, We have used a pretrained ResNet-50 network whose weights were not updated during the training of the generator and discrimiator. The input image and the target image are of the same class with different poses. The reconstruction loss (MSE) is incorporated with the negative discriminator loss to update the Generator. In our implementation, we have used 9 Residual blocks sequentially in the generator architecture. The discriminator is trained on the combined loss (binary crossentropy and categorical crossentropy).

The architecture of the proposed model is described in detail in section~\ref{methodology}. For training the Generator as well as Discriminator we have used Adam optimizer with $\beta_1=0.5$ and $\beta_2=0.999$. The initial learning rate was set to 0.0002 with a decay factor 10 at every 20 epoch. A batch size of 32 is taken as standard.
\section{Results and Discussion}
\subsection{Qualitative Results}
\begin{figure}[!ht]
	\centering
	\includegraphics[width=0.98\linewidth]{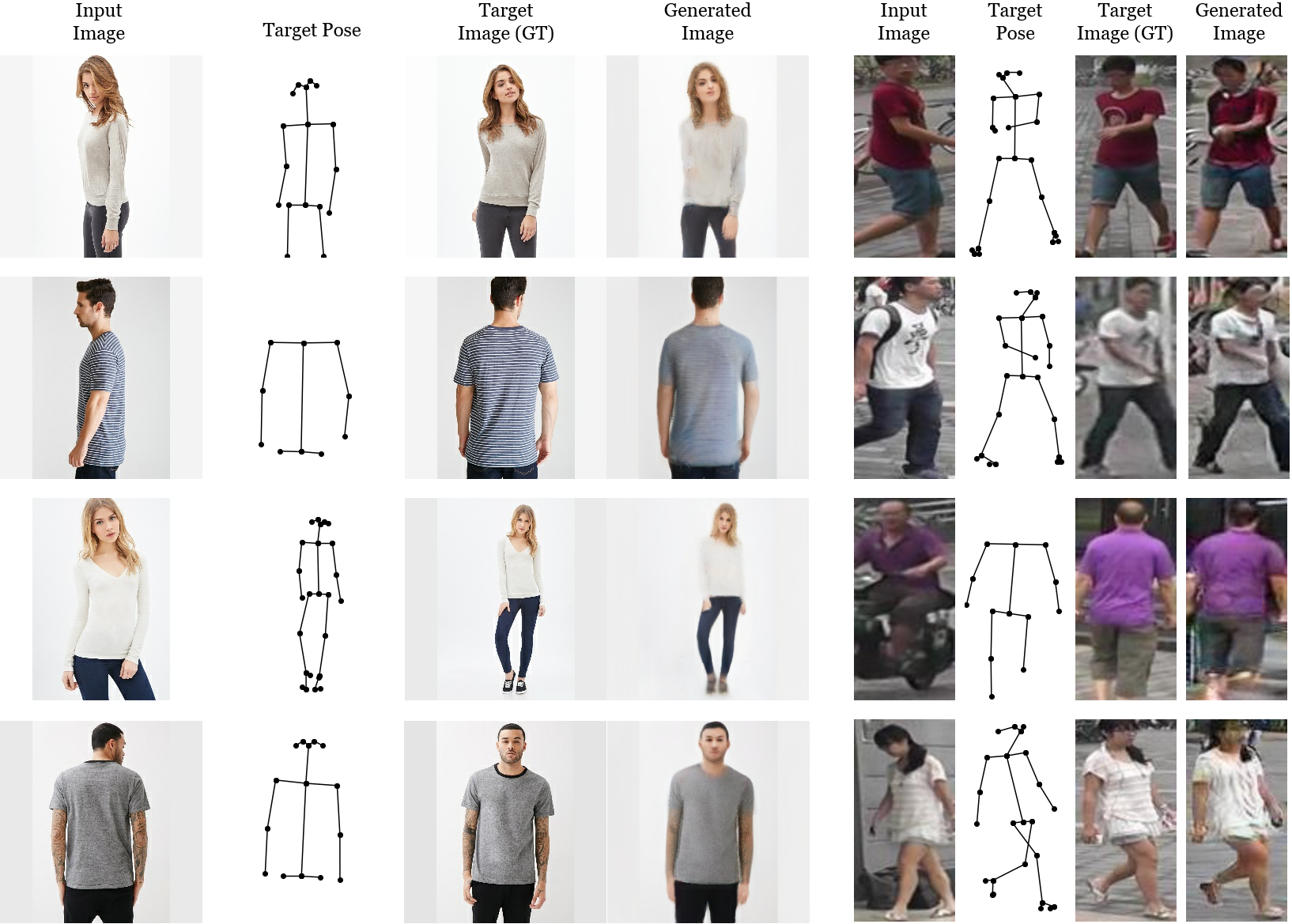}
	\caption{Qualitative results on DeepFashion and Market-1501 datasets. The proposed model is able to reproduce good details, and also learn invariable information like the colour and texture of clothing, characteristics of make/female attributes such as hair and face while successfully perform image transformation into the desired pose.}
	\label{fig:main_result}
\end{figure}

We demonstrate a series of results in high resolution fashion dataset DeepFashion~\cite{deepfashion} as well as a low resolution re-identification dataset Market-1501~\cite{market-1501}. In both the datasets, by visual inspection, we can say that our model performs good reconstruction and is able to learn invariable information like the colour and texture of clothing, characteristics of make/female attributes such as hair and face while successfully performing image transformation into the desired pose. The results on DeepFashion is better due to good details and simple background, whereas the low resolution affects the quality of the generated images in Market-1501. The results are demonstrated in Figure~\ref{fig:main_result}.
\vspace{-2mm}
\vspace{-8mm}
\subsection{Quantitative Results}
We use two popular measures of GAN performance, namely Structural Similarity (SSIM)~\cite{ssim} and Inception Score (IS)~\cite{is} for verifying the performance of our model. We compare our work with the already existing methods based on SSIM and IS scores on both DeepFashion and Market-1501 datasets in Table~\ref{tab:result-table}. Our model achieves the best IS score in Market-1501 dataset while achieving second best results in SSIM score in both the datasets. However, the deviation from the state-of-the-art is $\sim 1.5\%$ in these cases which can be overcome through rigorous testing and hyperparameter tuning. We also inspect the improvement incorporated by data augmentation as seen in Table~\ref{tab:result-table}. The proposed augmentation methods give an average improvement of $\sim 9\%$. This essentially strengthens our argument that a significant boost in performance can be gained by exploring effective training schemes, without changing the model parameters or loss function.
\begin{table}[!ht]
	\caption{Comparative study with existing methods in DeepFashion and Market-1501 datasets. The best and second best results are denoted in red and blue respectively.} \label{tab:result-table}
	\begin{center}
		\begin{tabular}{ C{4cm} C{1.2cm} C{1.2cm} C{1.2cm} C{1.2cm} }
			\hline \\[-2.5ex]
			{} & \multicolumn{2}{c}{DeepFashion} & \multicolumn{2}{c}{Market-1501} \\ \cline{2-5} \\[-2.4ex]
			Model & SSIM & IS & SSIM & IS \\  
			\hline \\[-2.5ex]			
			pix2pix~\cite{pix2pix} & 0.692 & 3.249 & 0.183 & 2.678\\
			PG2~\cite{pg2} & 0.762 & 3.090 & 0.253 & 3.460\\
			DSCF~\cite{deformable-gans} & 0.761 & \textcolor{red}{3.351} & 0.290 & 3.185\\
			BodyROI7~\cite{disentangled-pig} & 0.614 & 3.228 & 0.099 & \textcolor{blue}{3.483}\\
			Dong et al.~\cite{soft-gated} & \textcolor{red}{0.793} & \textcolor{blue}{3.314} & \textcolor{red}{0.356} & 3.409\\
			\hline \\[-2.5ex]
			Ours w/o augmentation & 0.713 & 3.006 & 0.268 & 3.425 \\
			Ours (full) & \textcolor{blue}{0.781} & 3.238 & \textcolor{blue}{0.302} & \textcolor{red}{3.488} \\
			\hline
		\end{tabular}
	\end{center}
\end{table}
\vspace{-8mm}
\subsection{Failure Cases}
\vspace{-4mm}
\begin{figure}[H]
	\centering
	\includegraphics[width=0.98\linewidth]{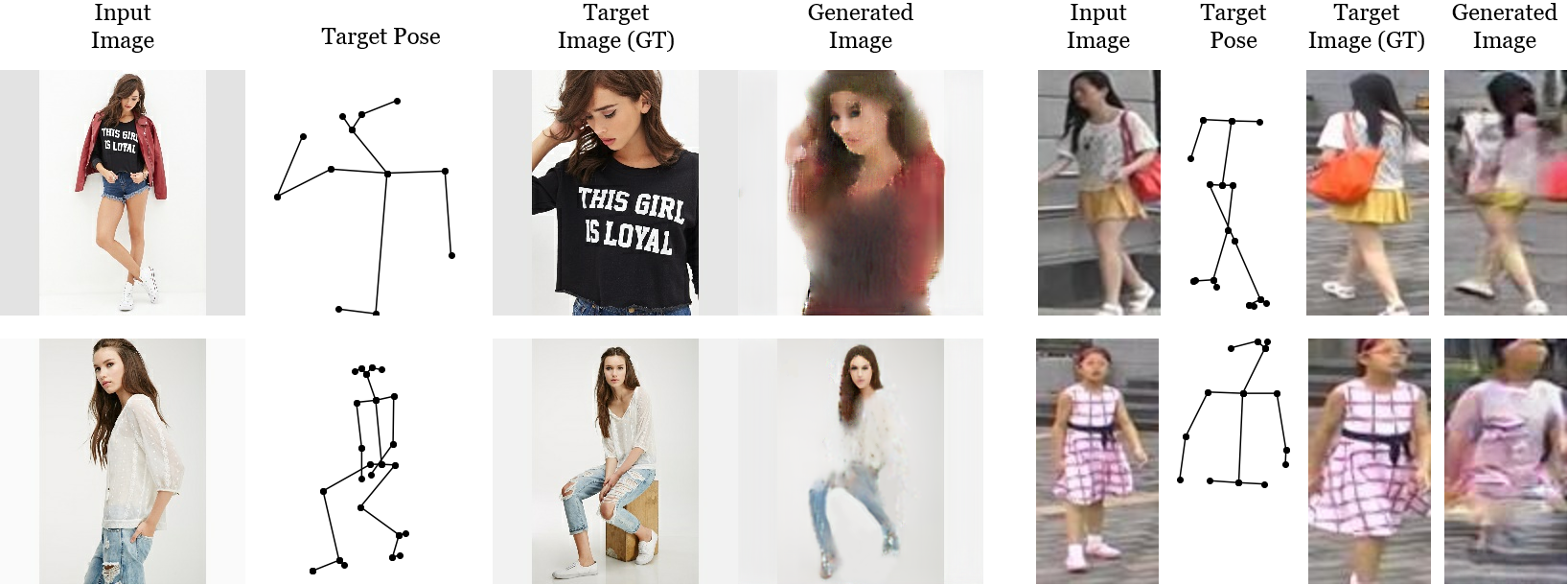}
	\caption{Failure Cases in our pt-GAN model. If the input contains fine details (text, stripes) or the target pose is incomplete then the reconstruction is poor. The external attribute (handbag) learning is of limited success.}
	\label{fig:fail_result}
\end{figure}
\vspace{-3mm}
We analyse some of our failure cases in both the datasets to understand the shortcoming of our model. As seen in Figure~\ref{fig:fail_result}, the text in clothing as well as very fine patterns of clothing (stripes, dots) are not modelled properly. The external attribute features (e.g. the handbag in Figure~\ref{fig:fail_result}) are not learned properly as it is difficult to map external attributes to the output image when conditioned only on pose. The accuracy is also dependent on the completeness of the target pose. Finally, there is some limitation in cases where a rare complex pose is presented which has scarce training data. In Market-1501, the reconstruction of faces is not very good due to poor resolution.
\subsection{Further Analysis}
\begin{figure}[!ht]
	\centering
	\includegraphics[width=0.6\linewidth]{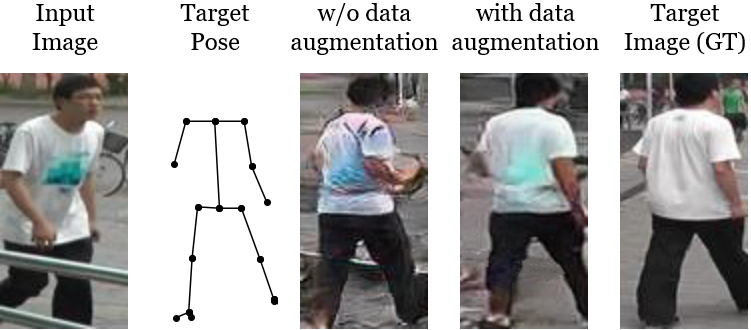}
	\caption{Occlusion invariance using the proposed model. The occlusion is partially carried forward when data augmentation methods are not used. With data augmentation, the resultant image is free from the artifacts.}
	\label{fig:occlusion_inv}
\end{figure}

Along with the quantitative and qualitative results, we demonstrate a special case to show the improvement caused by data augmentation methods. As seen in Figure~\ref{fig:occlusion_inv}, the occlusion in the input image is partially carried forward when the data augmentation methods are not used. With data augmentation the generated image is better in quality and the artifacts generated in the edges are less. 
\vspace{-6mm}
\section{Conclusion}
In this work, we proposed an improved end-to-end pose transformation model to synthesize photorealistic images of a given person in any desired pose without any external feature learning. We make use of the residual learning strategy with effective data augmentation techniques to achieve robustness in terms of occlusion, scale, illumination and distortion. For future work, we plan to achieve better results by utilising feature transport from the source image and conditioning the discriminator on both source image and target pose, alongwith using a perceptual (content) loss for reconstruction.
%
%
%
%
\bibliographystyle{splncs04}
\bibliography{mybibliography}
\end{document}